\documentclass[10pt,twocolumn,letterpaper]{article}

\usepackage{iccv}
\usepackage{times}
\usepackage{url}
\usepackage{epsfig}
\usepackage{graphicx}
\usepackage{amsmath}
\usepackage{enumitem}
\usepackage{amssymb}
\usepackage{mathrsfs}
\usepackage{bm}
\usepackage{booktabs}
\usepackage{subcaption}
\usepackage{multirow}
\usepackage{pifont}
\usepackage{arydshln}
\usepackage{color, xcolor}
\usepackage{colortbl, booktabs}
\usepackage{array}
\usepackage{soul}
\usepackage{lipsum}
\usepackage{marvosym}
\usepackage[accsupp]{axessibility}
\definecolor{light}{gray}{.85}
\definecolor{smooth}{gray}{0.95}
\definecolor{dg}{rgb}{0.0, 0.59, 0.09}
\usepackage[font=small,labelfont=bf]{caption}

\newcommand\blfootnote[1]{%
	\begingroup
	\renewcommand\thefootnote{}\footnote{#1}%
	\addtocounter{footnote}{-1}%
	\endgroup
}

\definecolor{linkColor}{rgb}{0.,0.44,0.74}
\usepackage[pagebackref=true,breaklinks=true,letterpaper=true,colorlinks,bookmarks=false,citecolor=linkColor]{hyperref}

\iccvfinalcopy 
\usepackage[capitalize]{cleveref}
\crefname{section}{Sec.}{Secs.}
\Crefname{section}{Section}{Sections}
\Crefname{table}{Table}{Tables}
\crefname{table}{Tab.}{Tabs.}

\def\iccvPaperID{3787} 

\ificcvfinal\fi

\begin{document}

\title{
Memory-and-Anticipation Transformer for Online Action Understanding
}

\author{
    Jiahao Wang$^{1*}$ \quad
    Guo Chen$^{1, 2*}$ \quad
    Yifei Huang$^{2}$ \quad
    Limin Wang$^{1, 2}$ \quad
    Tong Lu$^{1}$\textsuperscript{\Letter} \\
    $^1$ State Key Laboratory for Novel Software Technology, Nanjing University \\
    $^2$ Shanghai AI Laboratory \\
}

\maketitle
\thispagestyle{empty}
\blfootnote{* equal contribution, \Letter\ corresponding author (lutong@nju.edu.cn)}
\begin{abstract}
Most existing forecasting systems are memory-based methods, which attempt to mimic human forecasting ability by employing various memory mechanisms and have progressed in temporal modeling for memory dependency. Nevertheless, an obvious weakness of this paradigm is that it can only model limited historical dependence and can not transcend the past. In this paper, we rethink the temporal dependence of event evolution and propose a novel memory-anticipation-based paradigm to model an entire temporal structure, including the past, present, and future. Based on this idea, we present Memory-and-Anticipation Transformer (MAT), a memory-anticipation-based approach, to address the online action detection and anticipation tasks. In addition, owing to the inherent superiority of MAT, it can process online action detection and anticipation tasks in a unified manner. The proposed MAT model is tested on four challenging benchmarks TVSeries, THUMOS'14, HDD, and EPIC-Kitchens-100, for online action detection and anticipation tasks, and it significantly outperforms all existing methods.
Code is available at 
{\small \url{https://github.com/Echo0125/Memory-and-Anticipation-Transformer}}.
\end{abstract}
\section{Introduction}
\label{sec:intro}

Online anticipation~\cite{kitani2012activity} or detection~\cite{oad2016} in computer vision attempt to perceive future or present states from a historical perspective. They are the crucial factors of AI systems that engage with complex real environments and interact with other agents, \eg\ wearable devices~\cite{wearable-device}, human-robot interaction systems~\cite{human-robot-systems}, and autonomous vehicles~\cite{yu2020autonomous}. 
\begin{figure}
    \centering
    \includegraphics[width=0.48\textwidth]{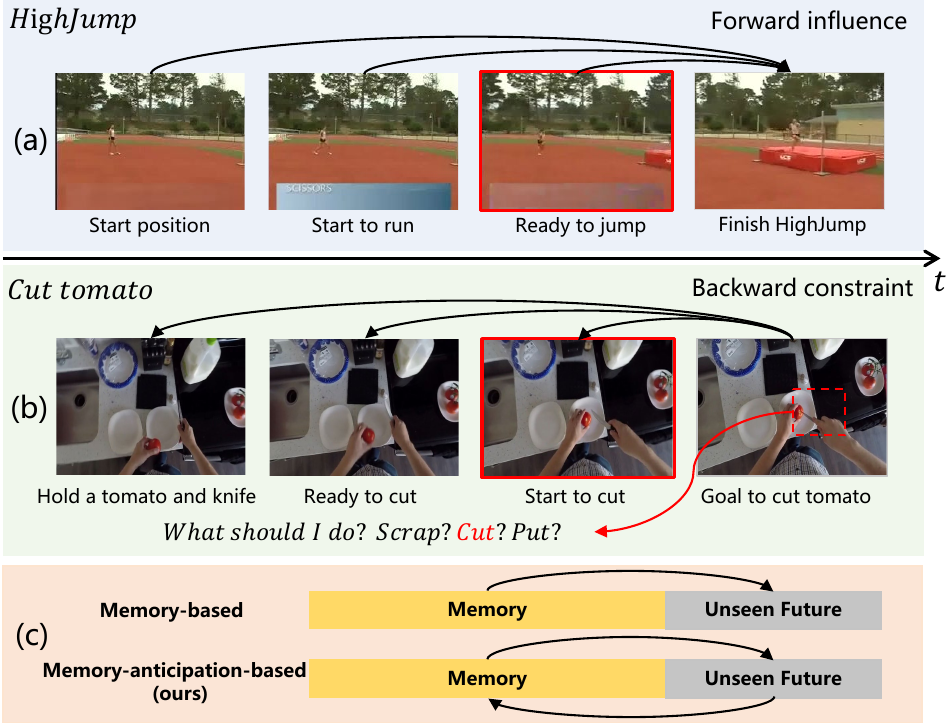}
    \caption{\textbf{Examples for the forward influence of memory and backward constraint of anticipation.} (a) An athlete prepares to start, starts to run, and then jumps to form the result of completing a high jump; (b) The goals of a camera wearer for cutting a tomato dictate the sequence of actions he will perform: hold the knife, get ready to cut, etc.; (c) Compared with the memory-based method, the memory-anticipation-based method is able to establish circular dependencies between memory and future.}
    \vspace{-3mm}
    \label{story}
\end{figure}

Human beings frequently imagine future events based on past experiences. Similarly, anticipating future events inherently requires modeling past actions or the progression of events to predict what will happen next.
To bridge the gap with human forecasting ability, current systems strive to mimic this cognitive ability by employing different memory mechanisms~\cite{lstr,huang2018predicting,gatehub,testra,tempagg,avt}. Previous research~\cite{rulstm,tempagg} demonstrated that modeling long temporal context is crucial for accurate anticipation. 
LSTR~\cite{lstr} further decomposes the memory encoder into long and short-term stages for online action detection and anticipation, allowing for the extraction of more representative memory features. 
Similarly, research efforts such as~\cite{testra,gatehub,gtrm} have also illustrated improvements based on this principle.

However, despite various \textbf{memory-based} approaches, memory is not the only driver of present or future action.
The synchronization between shot transitions and the evolution of actor behavior appears to cohere with the beliefs and wills of the event weavers, be they actors or camera-wearers. In (a) and (b) of Fig~\ref{story}, we use the examples of high jumping and cooking to show the forward influence of memory and backward constraint of anticipation.
Upon closer examination of both examples, it has been observed that future-oriented thoughts may impact action and memory, which seems modulated by the encoding of new information~\cite{2007ghosts,mcn}. 
Additionally, anticipation may change as ongoing behavior progresses and memory is updated~\cite{15active, 17neuroscience}. 
This indicates that there exists a circular interdependence between anticipation and memory,
constraining the evolution of behavior or events.

Based on the preceding analysis, we reevaluate the temporal dependencies inherent in \textbf{memory-based} methods. In their preoccupation with the impact of memory on anticipation, these memory-based methods tend to overlook the inverse direction of impact, \ie, anticipation on memory. Consequently, historical representations are not adequately corrected, thereby risking impeding any attempts to transcend the past. Against this backdrop, we assert that a comprehensive temporal structure seamlessly integrates memory and anticipation is indispensable. In other words, a \textbf{memory-anticipation-based} method building circular dependencies between memory and anticipation, as shown in Fig~\ref{story}(c), holds tremendous promise for enhancing cognitive inference capabilities and advanced understanding of AI systems about the present and future.

To this end, we propose \textbf{M}emory-and-\textbf{A}nticipation \textbf{T}ransformer \textbf{(MAT)}, a novel \textbf{memory-anticipation-based} approach that fully
models the complete temporal context, including history, present, and future.
A \emph{Progressive Memory Encoder} is designed to provide a more precise history summary by compressing long- and short-term memory in a segment-based fashion.
Meanwhile, we propose our key idea of modeling circular dependencies between memory and future, implemented as  \emph{Memory-Anticipation Circular Decoder}. It first \emph{learns latent future features} in a supervised manner, then updates iteratively the enhanced short-term memory and the latent future features by performing \emph{Conditional circular Interaction} between them. Among them, multiple interaction processes capture the circular dependency and supervise the output to maintain stable features with real semantics.

Remarkably, owing to the inherent superiority of our model design, we are able to adapt both tasks, \ie, online action detection and anticipation, in a unified manner, spanning the training and online inference stages. For any given dataset, the MAT model obviates the need for separate training or testing for each task. Rather, a single training process is enough, and during inference, the corresponding token for each task can be extracted effortlessly.

In summary, our contributions are 1). We rethink the temporal dependence of event evolution and propose a memory-anticipation-based paradigm for the circular interaction of memory and anticipation, introducing the concept of memory and future circular dependence.
2). We propose a unified architecture Memory-Anticipation Transformer (MAT) that simultaneously processes online action detection and anticipation, showing effective performance. 3). MAT significantly outperforms all existing methods on four challenging benchmarks for online action detection and anticipation tasks, \ie, TVSeries~\cite{oad2016}, THUMOS'14~\cite{THUMOS14}, HDD~\cite{hdd} and EPIC-Kitchens-100~\cite{ek100}.

\section{Related Work}
\label{sec:work}
\textbf{Online Action Detection and Anticipation.}
Online action detection~\cite{oad2016} aims to predict the action as it happens without accessing the future. Previous methods~\cite{trn, idn, woad, colar, oadtr, lstr, gatehub, testra, mt} for online action detection include recurrent networks, reinforcement learning, and, more recently, transformers. 
\cite{red} uses a reinforced network to encourage making great decisions as early as possible. \cite{trn} adopts RNN to generate future predictions to improve online action detection.
\cite{lstr} proposes a transformer to scale the history spanning a longer duration.
The target of action anticipation~\cite{ek100, rulstm, avt, memvit} is predicting what actions will occur after a certain time. RULSTM~\cite{rulstm} explores action anticipation and proposes an LSTM-based network to solve this problem. AVT~\cite{avt} proposed an end-to-end attention-based model for anticipative video modeling and learns feature representations by self-supervised methods.

\begin{figure*}
    \centering
    \includegraphics[width=1\textwidth]{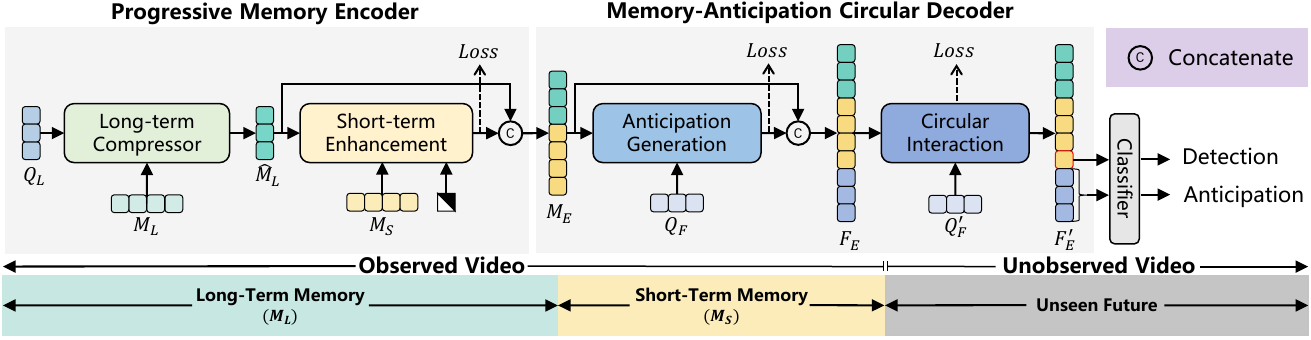}
    \vspace{-2em}
    \caption{\textbf{MAT architecture.} We first divide the temporal sequence into long-term and short-term memories. 
    In \emph{Progessive Memory Encoder}, long-term memory queries $\mathbf{Q}_{L}$ progressively map the long-term to an abstract representation and be fed to the transformer decoder block to enhance the short-term memory. Then the \emph{Memory-Anticipation Circular Decoder} utilizes learnable queries $\mathbf{Q}_{F}$ and $\mathbf{Q}^{\prime}_{F}$ to perceive the future context and circularly updates the historical and future representation. Finally, a weight-shared classifier is adopted to output the classification scores of short-term and future for online action detection and anticipation.}
    \label{fig:framework}
\end{figure*}

Online action detection and anticipation share two key characteristics: the inability to perceive future representations, with information derived solely from historical context, and the goal of predicting actions after a certain time. 
Previous research has typically studied the model for each task in isolation~\cite{gatehub, colar} or trained and tested them separately~\cite{lstr,testra}, ignoring the potential benefits of leveraging their shared characteristics. We propose a novel unified framework, named MAT, to address this limitation, which allows us to train or infer both tasks simultaneously.

\textbf{Transformer for Video Understanding.} Transformers have proven successful in natural language processing \cite{transformer}, and researchers have shown increasing interest in its application to vision tasks \cite{vit, deit, swin, pvt, omnivore, li2023fb}. Several methods have been proposed to use transformers for temporal modeling in video-related tasks~\cite{vivit, longformer, videollm, cpmt, li2022uniformerv2, li2023unmasked, TadTR, linformer, wang2022internvideo, vidtr}.
Timesformer~\cite{timesformer} proposes divided attention to capture spatial and temporal information for action recognition separately. 
MViT~\cite{mvit} builds a feature hierarchy by progressively expanding channel capacity and reducing video spatial-temporal resolution. Actionformer~\cite{actionformer} adopts the transformer architecture with local attention for temporal action localization. 
TubeDETR~\cite{tubedetr} tackles video grounding with an encoder-decoder transformer-based architecture to efficiently encode spatial and multi-modal interactions.
Our MAT model is also built upon the transformer, utilizing its encoder and decoder architectures. We meticulously integrate the transformer blocks to implement our memory encoder and memory-anticipation circular decoder.

\section{Problem Setup}
\label{sec:problem}
Online action anticipation and detection share a similar goal of predicting actions in untrimmed video sequences. In the former, we aim to forecast actions after a time gap $\tau$, as defined in~\cite{ek100}, while in the latter, we predict the action under $\tau=0$.
Both tasks require classifying actions without access to future information during inference.
We represent the input video as 
$ \mathbf{V} = \{ v_t \}_{t = -T + 1}^{0} $,
and our objective is to predict the action category $\hat{y}_{\tau} \in \{ 0, 1, 2, \dots, C \}$ occurs after a specified time gap $\tau$ in the future. Here, $C$ represents the total number of action categories, and label $ 0 $ denotes the background category.

\section{Memory-and-Anticipation Transformer}
\label{sec:approach}
We now describe Memory-and-Anticipation Transformer (MAT) model architecture, as illustrated in Fig~\ref{fig:framework}. It processes the memory cached in runtime to predict current or future actions. We adopt the \emph{encoder-decoder} architecture to implement the model. The \emph{Progressive Memory Encoder} is proposed to improve the quality of compressing long-term memory and enhancing short-term memory, which has a fundamental impact on anticipation and detection. MAT also employs our presented key component \emph{Memory-Anticipation Circular Decoder} that generates the latent anticipation feature and conducts interaction between memory and anticipation in an iterative loop.
We now introduce each model component in detail, followed by the training, online inference, and implementation details.

\subsection{Progressive Memory Encoder}

We use video feature $\mathbf{F} = \{ f_t \}_{t = -T + 1}^{0} \in \mathbb{R}^{T \times D}$ generated by a pretrained feature extractor as the input to our model, where $D$ and $T$ are the dimension and length of the feature sequence, respectively. 
To better handle the long memory, following~\cite{lstr}, we divide the feature sequence into two consecutive memories. The first is the short-term memory that stores only a handful of recently occurred frames. It stores the feature vectors as $ \mathbf{M}_S = \{ f_t \}_{t = -m_S + 1}^{0}$, where $ m_S $ denotes the length. The other, named long-term memory, contains the feature that is far away from the current time, which is defined as $\mathbf{M}_L = \{ f_t \}_{t = -T + 1}^{-m_S} $. We set $ m_L $ to be the length of the long-term memory, which is much longer than the short-term memory.

In practice, \emph{Progressive Memory Encoder} progressively encodes long- and short-term memories. It first compresses the long-term memory to an abstract representation and injects them into the short-term memory. 
Differing with previous works~\cite{lstr, testra} that separately tackle the long- and short-term memories by one long-term encoder and one short-term decoder, we integrate these two works into a single encoder architecture.
It is due to our encoded short-term memory being served for decoders and not as the final output. In our experiments, we observe that the design of the memory encoder is crucial for model performance. Therefore, different with~\cite{lstr, testra}, we introduce a novel long-term memory compression method to improve the quality of memory encoding.

\begin{figure}
    \centering
    \includegraphics[width=0.48\textwidth]{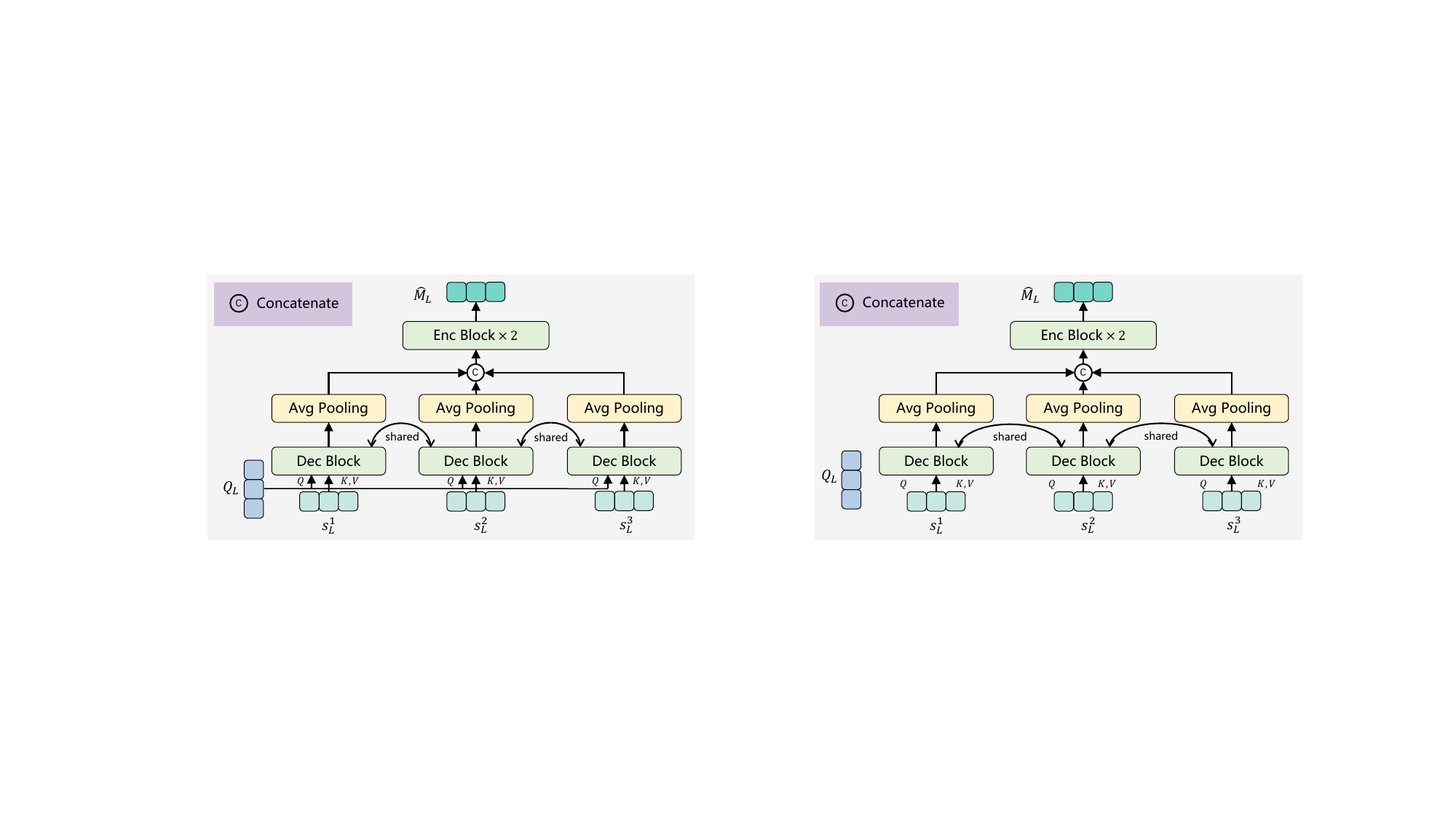}
    \caption{\textbf{Segment-based Long-term Memory Compression.} 
    For example, given $ m_L = 9, N_s = 3$, the process is shown above. We can finally obtain compressed features $ \widehat{M}_L $ of length 3.}
    \label{fig:encoder}
\end{figure}

\textbf{Segment-based Long-term Memory Compression.}
LSTR~\cite{lstr} adopts a two-stage compression technique to project the long-term memory to a fixed-length latent embedding. It implements a function similar to the bottleneck layer~\cite{resnet}. 
Long-term memory, however, represents a long token sequence that often carries much noise.
It is problematic to generate attention weights while cross-attention queries too many tokens with mixed noise. 
We propose \emph{Segment-based Long-term Memory Compression} to alleviate the issue, as shown in Fig~\ref{fig:encoder}.
Concretely, we first divide $\mathbf{M}_{L}$ into $N_s$ non-overlapping memory segments  
$ \mathbf{S} = \{ \mathbf{s}_{i}\}^{N_s}_{i=1} $ of length $\frac{m_L}{N_s}$.
We then employ $N_L$ learnable tokens $ \mathbf{Q}_{L} \in \mathbb{R}^{ N_L \times D }$ as the long-term memory queries and a weight-shared transformer decoder block to query each segment. 
$\mathbf{S}$ will be transformed to $N_s$ segment-level abstract feature $\mathbf{F} =\{ \mathbf{f}_{i}\}^{N_s}_{i=1}$, where $\mathbf{f}_{i} \in \mathbb{R}^{ N_L \times D }$. We average pool each $\mathbf{f}_{i}$ to one vector $\mathbb{R}^{ D }$ and concatenate them to form the long-term compressed segmented memory $\mathbf{M}_L^s \in \mathbb{R}^{ N_s \times D }$.
At last, we feed them into two transformer encoder blocks to obtain the final compressed long-term memory $ \mathbf{\widehat{M}}_L \in \mathbb{R}^{N_s \times D} $.

\textbf{Short-term Memory Enhancement.}
As short-term memory $ \mathbf{M}_S = \{ f_t \}_{t = -m_S + 1}^{0} $ contains crucial trends for accurately predicting the current or upcoming activities, we leverage short-term memory as the query to retrieve relevant context from the compressed long-term memory.
As illustrated in Fig~\ref{fig:framework}, we employ a transformer causal decoder block to aggregate compressed long-term memory $\mathbf{\widehat{M}}_L$ into short-term memory. The enhanced short-term memory $\mathbf{\widehat{M}}_S$ is then used to identify actions at each timestep, constituting a sequence of previously occurring actions and predicting future states and actions. We use frame-level action labels to supervise the dense classification task, encouraging the model to produce a clear sequence of historical behavior.

\subsection{Memory-Anticipation Circular Decoder}
\label{decoder}
To implement our key idea that considers the co-reaction between anticipation and memory, we carefully design \emph{Memory-Anticipation Circular Decoder} that circularly performs interaction between them. It first generates the latent anticipation that is a prerequisite for memory-anticipation interaction. Then, \emph{Conditional Circular Interaction} will generate new memories and anticipation conditioned on old ones iteratively. Finally, a weight-shared classification head will supervise the generated memories and anticipation.

\textbf{Latent Anticipation Generation.}
Initially, we do not have an anticipation state. Therefore, we must first generate a latent anticipation feature, \ie, a future representation containing abstract information derived from known memory. 
Given the compressed long-term memory $\mathbf{\widehat{M}}_L$ and short-term memory $\mathbf{\widehat{M}}_S$, we first concatenate them to form the entire memory $\mathbf{M}_E = [\mathbf{\widehat{M}}_L, \mathbf{\widehat{M}}_S]  $, where $[\cdot,\cdot]$ is the concatenating operation along the temporal dimension. 
Then, we use $N_F$ future queries $\mathbf{Q}_F \in \mathbb{R}^{ N_F \times D }$ and a transformer decoder block to query $\mathbf{M}_E$. The updated future queries are represented as the latent anticipation features $\mathbf{F}_{A}\in \mathbb{R}^{ N_F \times D}$ at the current state. We use future information, \eg, future video features and future action labels, to supervise $\mathbf{F}_{A}$. In practice, we use information from the next $T_F$ seconds. Since $\mathbf{F}_{A}$ is only used to describe the latent future states, too large future embeddings may overfit the anticipation representation. We set $N_F$ to be smaller than the length of the used future information sequence. Then we use bilinear interpolation to upsample it to align for point-to-point supervision. Note that although we use future information here, this does not lead to an information leak. The model will generate preliminary anticipation in the inference phase and not use future labels or features.

\textbf{Conditional Circular Interaction.}
So far, we have generated long-term memory $\mathbf{\widehat{M}}_L$, short-term memory $\mathbf{\widehat{M}}_S$ and latent anticipation features $\mathbf{F}_A$. The memories, however, are limited to the past and can not go beyond history. 
We now describe how \emph{Conditional Circular Interaction} works.
For brevity, we detail how memory and anticipation interact with each other one time.

\begin{figure}
    \centering
    \includegraphics[width=0.48\textwidth]{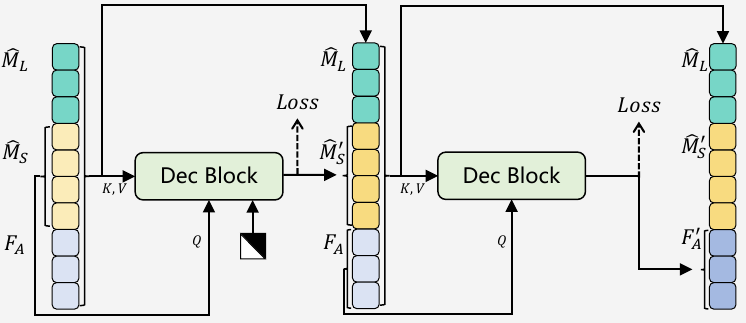}
    \caption{\textbf{Conditional circular interaction} dynamically updates and re-constructs the context between memories and anticipation by aggregating different temporal information. }
    \label{cci}
\end{figure}

As shown in Fig~\ref{cci}, we first concatenate $\mathbf{\widehat{M}}_L$, $\mathbf{\widehat{M}}_S$ and $\mathbf{F}_A$ to form the entire feature $\mathbf{F}_E = [\mathbf{\widehat{M}}_L, \mathbf{\widehat{M}}_S, \mathbf{F}_A]$. 
Then, $\mathbf{F}_E$ and $\mathbf{\widehat{M}}_S$ are feed into a transformer decoder block:
\begin{equation}
    \begin{aligned}
        \mathbf{\widehat{M}}_S^{\prime} = \text{CrossAttn}(\mathbf{\widehat{M}}_S,\mathbf{F}_E,\mathbf{F}_E),
    \end{aligned}
\end{equation}
where $\text{CrossAttn}(Q,K,V)$ is the cross-attention layer in the transformer decoder block. The updated short-term memory is represented as $\mathbf{\widehat{M}}_S^{\prime}$. Since $\mathbf{F}_E$ contains memories and anticipation, $\mathbf{\widehat{M}}_S$ as a query condition dynamically extracts new semantics mixed with memories and anticipation from three representations.  Next, we re-concatenate $\mathbf{\widehat{M}}_L$, $\mathbf{\widehat{M}}_S^{\prime}$ and $\mathbf{F}_A$ to generate the new entire feature $\mathbf{F}_E^{\prime} = [\mathbf{\widehat{M}}_L, \mathbf{\widehat{M}}_S^{\prime}, \mathbf{F}_A]$. Another transformer decoder block is used to query $\mathbf{F}_E^{\prime}$, with $\mathbf{F}_A$ as the query condition:
\begin{equation}
    \begin{aligned}
        \mathbf{\mathbf{F}}_A^{\prime} = \text{CrossAttn}(\mathbf{F}_A,\mathbf{F}_E^{\prime},\mathbf{F}_E^{\prime}),
    \end{aligned}
\end{equation}
where $\mathbf{F}_A^{\prime}$ is the resulting anticipation features.
After that, $\mathbf{F}_A^{\prime}$ is injected with the higher-order inferred semantics brought about by the interaction between memory and anticipation.

In the interaction process, the generated $\mathbf{\widehat{M}}_S^{\prime}$ is both supervised like the above initial anticipation features. It should be noted that different from generating latent initial anticipation through fewer future queries, we find that using a new group of future queries $\mathbf{Q}_F^{\prime} \in \mathbb{R}^{ N_F^{\prime} \times D }$ strictly aligned future information sequences of $T_F$ seconds in the first interaction process to replace the $\mathbf{F}_A$ with $\mathbf{Q}_F^{\prime}$, bring performance improvement. Thus, we point-to-point supervise the anticipation $\mathbf{F}_A^{\prime}$ generated by each interaction process. Meanwhile, this design is more natural to output the corresponding token according to the anticipation gap time.
We circularly perform the interaction process $N_t$ times until saturation. The yielded  $\mathbf{\widehat{M}}_S^{\prime}$ and $\mathbf{F}_A^{\prime}$ in one interaction process become the new $\mathbf{\widehat{M}}_S$ and $\mathbf{F}_A$ and are passed to the next interaction process. 

\subsection{Training}
\label{training}
Our MAT model relies on action steps of short-term memory to predict the current or future action. 
These steps are utilized as auxiliary information to benefit action detection and anticipation. 
It is worth noting that since the MAT model regards online action detection and anticipation as the same tasks with different settings, we can define the loss function in a unified fashion.

For the  $\mathbf{\widehat{M}}_S$ generated by the encoder, we feed it to a classifier to generate the action probabilities $ \mathbf{\widehat{Y}}_S = \{ \mathbf{\widehat{y}}_S^{i} \}_{i=1}^{m_S} $. 
We then supervise the whole short-term memory using a cross-entropy loss with the labeled action, $ \{ c_S^{i} \}_{i=1}^{m_S} $. 
Learned future features $ \mathbf{F}_A $ are fed into the same classifier to generate the probabilities $ \mathbf{\widehat{Y}}_F = \{ \mathbf{\widehat{y}}_F^{i} \}_{i=1}^{N_F} $ and we also adopt a cross-entropy loss with the target, $ \{ c_F^{i} \}_{i=1}^{m_S} $:
\begin{equation}
    \begin{aligned}
        \mathcal{L}_S^{0}  = -\sum_{i=1}^{m_S}\log\mathbf{\widehat{y}}_S^{i}[c_S^i],
        \mathcal{L}_F^{0}  = -\sum_{i=1}^{N_F}\log\mathbf{\widehat{y}}_F^{i}[c_F^{i}],
    \end{aligned}
\end{equation}
where $\mathcal{L}_S^{0}$ and $\mathcal{L}_F^{0}$ are the loss function of encoded short-term memory and initial anticipation features, respectively.
As the conditional circular interaction is conducted $ N_t $ times, we use $\mathcal{L}_S^{i}$ and $\mathcal{L}_F^{i}$ to represent the loss of memory and anticipation in each interaction process, where $1 \leq i \leq N_t$. Their calculation is similar to $\mathcal{L}_S^{0}$ and $\mathcal{L}_F^{0}$.
The final training loss is formulated by:
\begin{equation}
    \begin{aligned}
        \mathbf{\mathcal{L}}=\sum_{i=0}^{N_t}\lambda_{s}^{i}\cdot\mathcal{L}_S^{i}+\lambda_f\sum_{i=0}^{N_t}\mathcal{L}_F^{i},
    \end{aligned}
\end{equation}
where $ \lambda_s^i $ and $ \lambda_f $ are the balance coefficients.

\begin{table*}[t]
    \centering
    \small
    
    \subfloat[\textbf{Segment number.}\label{tab:segment}]{
    \setlength{\tabcolsep}{2mm}{
    \begin{tabular}[t]{cc}
        $ N_s $ & mAP \\
        \midrule
        ~\cite{lstr}& 69.6 \\
        2 &  69.8 \\
        4 & 70.2 \\
        \rowcolor{light}
        8 & \textbf{70.4} \\
        12 &  70.1 \\
    \end{tabular}}}
    \hspace{1.5mm}
    \subfloat[\textbf{Future queries renewing.}\label{tab:renew}]{
    \setlength{\tabcolsep}{2mm}{
    \begin{tabular}[t]{ccc}
    renewal & mAP \\
    \midrule
    0 &  70.0 \\
    \rowcolor{light}
    1 & \textbf{70.4}\\
    2 & 69.7\\
    \\
    \\
    \end{tabular}}}
    \hspace{1.5mm}
    \subfloat[\textbf{Future supervision.}\label{tab:future supervision}]{
    \setlength{\tabcolsep}{1.5mm}{
    \begin{tabular}[t]{ccc}
        deep & supervision & mAP \\
         \midrule
         - & cls & 69.9 \\
         \checkmark & feat & 69.6 \\
         \rowcolor{light}
         \checkmark  & cls & \textbf{70.4} \\
         \checkmark  & feat + cls &  70.2 \\
    \\
    \end{tabular}}}
    \hspace{1.5mm}
    \subfloat[\textbf{Shared classifier.}\label{tab:classifier}]{
    \setlength{\tabcolsep}{1.5mm}{
    \begin{tabular}[t]{ccc}
        short & future & mAP \\
         \midrule
          \multicolumn{2}{c}{unshared} & 69.6 \\
          \multicolumn{2}{c}{separately shared} & 70.1 \\
         \rowcolor{light}
          \multicolumn{2}{c}{full shared} & \textbf{70.4}\\
          \\
         
    \\
    \end{tabular}}}
    \hspace{1.5mm}
    \subfloat[\textbf{MixClip+.}\label{tab:mc+}]{
    \setlength{\tabcolsep}{1mm}{
    \begin{tabular}[t]{llccc}
        long & short  & V & N & A  \\
        \midrule 
        - & - & 32.1 & 36.1 & 18.1 \\
        MC & - & 32.9 & 36.9 & 18.4 \\
        MC & MC & 31.0 & 37.4 & 18.5 \\
        \rowcolor{light}
        MC & MC+ & \textbf{35.0} & \textbf{38.8} & \textbf{19.5}\\
        MC+ & MC+ & 31.6 & 36.8 & 18.4 \\
    \end{tabular}}}
    \caption{\textbf{Ablation Experiments.} We conduct detailed ablation on (a): Segment number, (b): Future queries renewing, (c): Future supervision, (d): Shared classifier, and (e): MixClip+. The \sethlcolor{light}\hl{gray rows} denote default choices.}
\end{table*}

\subsection{Online Inference}
The existing works~\cite{trn, colar, lstr, testra, avt} test each task independently. 
Thanks to the design of the unified framework, MAT simultaneously infers online action detection and anticipation tasks on one streaming video. It uses the output  $\mathbf{\widehat{M}}_S$ and $ \mathbf{F}_A$ produced by the last interaction as results of online action detection and anticipation, respectively. We take out the last token of $\mathbf{\widehat{M}}_S$ for online action detection. For action anticipation, according to gap time $ \tau$, we take out the corresponding token from $ \mathbf{F}_A$ for forecasting.

\section{Experiments}
\label{sec:experiments}

\subsection{Dataset and Metrics}
\textbf{Dataset.} We test on three online action detection datasets. \emph{THUMOS'14}~\cite{THUMOS14} consists of over 20 hours of sports video annotated with 20 actions. \emph{TVSeries}~\cite{oad2016} includes 27 episodes of 6 popular TV series, about 150 minutes each and 16 hours total. \emph{HDD}~\cite{hdd} is a large-scale human driving video dataset comprising 104 hours of untrimmed videos and 11 action categories. Furthermore, we evaluate on \emph{EpicKitchens-100}~\cite{ek100} that contains 100 hours of egocentric videos with at least 90K action segments and whose narrations are mapped to 97 verb classes and 300 noun classes.

\textbf{Evaluation Metrics.} For online action detection, following previous works \cite{oad2016, LAP-Net, oadtr, lstr, colar}, we use per-frame \emph{mean Average Precision (mAP)} on THUMOS'14 and HDD, and per-frame \emph{mean calibrated Average Precision (mcAP)}~\cite{oad2016} on TVSeries. For action anticipation, we employ \emph{Top-5 Verb/Noun/Action Recall}  to measure the performance with an anticipation period $\tau = 1s$~\cite{ek100}.

\subsection{Implementation details} 
Following prior works~\cite{trn, lstr, testra}, we conduct our experiments on pre-extracted features. For TVSeries and THUMOS'14, we first resample the videos at 24 FPS and then extract the frames at 4 FPS for training and validation. We adopt a two-stream network~\cite{tsn} to extract feature.  
We use off-the-shell checkpoint released by mmaction2~\cite{mmaction2} that pretrained on ActivityNet v1.3 \cite{anet} and Kinetics \cite{kinetics} to extract frame-level RGB and optical features. On EK100, following~\cite{rulstm}, we resample the videos at 30 FPS and then fine-tune the two-stream TSN on the classification task.

\textbf{Memory Dropping.}
Despite the promising capabilities of the dense attention mechanism, capturing long-range dependencies remains a challenge.
However, relying on the dense attention mechanism may lead to sub-optimization of the network, resulting in a locally optimal solution. 
We seek to explore memory-dropping strategies for implementing sparse attention during training.
We experiment with various approaches, including dropout~\cite{dropout}, top-k selection~\cite{knn}, and token dropping~\cite{mae}. We compare these strategies, recorded in the supplementary material, and find that the top-k selection yields the best performance.

\begin{table}[t]
    \centering
    \small
    \setlength{\tabcolsep}{1.0mm}{
    \begin{tabular}{cccccccccc}
     &long & short & future & interaction & $N_t$ & detection & anticipation \\
    \midrule
    \rowcolor{smooth}
    (a) &  &  &  & - & - & 67.6$_{\textcolor{dg}{+0.0}}$ & 53.7$_{\textcolor{dg}{+0.0}}$ \\
    (b)  & & \checkmark &  & CA & 1 &  68.3$_{\textcolor{dg}{+0.7}}$ & 54.7$_{\textcolor{dg}{+1.0}}$\\
   (c) & \checkmark & \checkmark &  & CA & 1 & 68.9$_{\textcolor{dg}{+1.3}}$  & 55.3$_{\textcolor{dg}{+1.6}}$ \\
    (d) &\checkmark &  & \checkmark & CA & 1 & 68.6$_{\textcolor{dg}{+1.0}}$ & 55.8$_{\textcolor{dg}{+2.1}}$\\
    (e) & & \checkmark & \checkmark & CA & 1 & 69.2$_{\textcolor{dg}{+1.6}}$ & 56.1$_{\textcolor{dg}{+2.4}}$\\
    \hline
   (f) & \checkmark & \checkmark & \checkmark & Avg\&Cat & 1 & 68.3$_{\textcolor{dg}{+0.7}}$ & 55.3$_{\textcolor{dg}{+1.6}}$\\
  (g)& \checkmark&\checkmark &\checkmark & CA & 1 &  69.9$_{\textcolor{dg}{+2.3}}$ & 56.6$_{\textcolor{dg}{+2.9}}$\\
    \rowcolor{light}
  (h) &\checkmark &\checkmark &\checkmark & CA & 2 & \textbf{70.4$\bf{_{\textcolor{dg}{+2.9}}}$} & \textbf{57.3$_{\textcolor{dg}{+3.6}}$} \\
  (i) &\checkmark &\checkmark &\checkmark & CA & 3 & 70.2$_{\textcolor{dg}{+2.6}}$ &  57.1$_{\textcolor{dg}{+3.4}}$ \\
    \end{tabular}}
    \caption{\textbf{Ablation study} on the information stream, interaction method, and times for Conditional Circular Interaction. The \sethlcolor{smooth}\hl{shallow gray row} denotes the baseline setting without any interaction.}
    \vspace{-3mm}
    \label{tab: interaction}
\end{table}

\textbf{MixClip+.}
TesTra~\cite{testra} proposed an augmentation technique called MixClip to solve over-fitting caused by large long-term memory. 
Unlike long-term memory which presents an abstract concept, short-term memory provides a continuous motion trend, often playing a decisive role in anticipation. It leads to the fact that MixClip damages the continuity of motion, thereby, does not apply to short-term memory. We propose MixClip+ to solve the issue. Due to space limitations, we will describe this part in the supplementary material.

In addition to mixing augmentation for long-term memory, we mix each feature token of short-term memory with a random clip from different videos. To maintain the continuity of short-term memory, we adopt soft fusion with hyper-parameter $\alpha = 0.25$ to mix features and the corresponding labels, similar to Mixup~\cite{mixup}.

\subsection{Ablation Study}
\label{sec:ablation}

Now we analyze the MAT model, using the two-stream features pretrained on ActivityNet v1.3 and THUMOS'14 test set as the test bed. Furthermore, we also study the contribution of the proposed MixClip+ on EPIC-Kitchens-100.

\textbf{Segment Number.}
We evaluate the effect of segment number $N_s$ for long-term memory compression in Table~\ref{tab:segment}. We find even a small $N_s$ can improve performance, compared with two-stage compression~\cite{lstr}. However, too many segments result in too little information within each segment to effectively capture temporal dependency. We use $ N_s = 8$ as the default in the following ablation experiments.

\begin{figure}[t]
    \centering
    \includegraphics[width=0.45\textwidth]{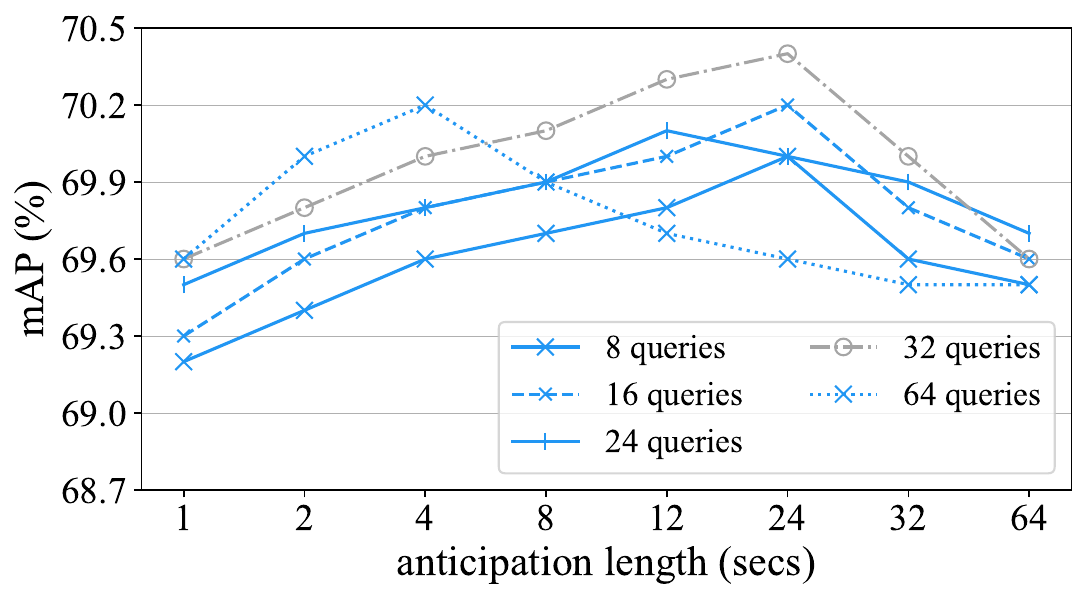}
    
    \caption{\textbf{Generating latent anticipation} with different anticipation length $T_F$ and numbers of learnable tokens $N_F$.}
    \vspace{-3mm}
    \label{fig:query_length}
\end{figure}

\textbf{Query Number and Anticipation Length.}
Fig~\ref{fig:query_length} compares generating latent anticipation using $N_F$ learnable tokens with different anticipation lengths $T_F$. The experimental results indicate that, for $T_F \leq 4$, a larger $N_F$ yields superior results owing to strongly correlated semantics of short-term memory and short-term anticipation thus, the model can benefit from more tokens.
We observe that $N_F=8, 16, 32$ exhibits a similar trend in the accuracy curve, with the model achieving optimal performance with $N_F=32$. This suggests that long-term anticipation is more adaptable for sparse tokens. The observations imply that anticipation is also suitable for long- and short-term modeling.

\textbf{Future Queries Renewal.}
As Section~\ref{decoder} outlines, renewing future queries before the initial interaction improves performance. In Table~\ref{tab:renew}, we compare different renewing times (``0'' denotes no renewal and utilizes $F_A$). A plausible explanation is that, in the first interaction process, the interaction from anticipation to memory adjusts the old memory, causing its domain to shift largely, which leads to insufficient interaction from the new memory to the initial anticipation feature.

\begin{table*}[t]
    \centering
    \small
    \setlength{\tabcolsep}{2.5mm}{
    \begin{tabular}{lcccc>{\columncolor{light}}ccccccc}
        \multirow{2}{*}{Method} & \multirow{2}{*}{Modality} & \multirow{2}{*}{Init} & \multicolumn{3}{c}{Overall} & \multicolumn{3}{c}{Unseen} & \multicolumn{3}{c}{Tail} \\
        \cmidrule(lr){4-6} \cmidrule(lr){7-9} \cmidrule(lr){10-12}
         & & & Verb & Noun & \cellcolor{white}Action & Verb & Noun & Action & Verb & Noun & Action \\
         \midrule
         RULSTM~\cite{rulstm} & \multirow{6}{*}{RGB} & IN-1K & 27.5 & 29.0 & 13.3 & 29.8 & 23.8 & 13.1 & 19.9 & 21.4 & 10.6 \\
         AVT~\cite{avt} &  & IN-1K & 27.2 & 30.7 & 13.6 & - & - & - & - & - & - \\
         AVT~\cite{avt} &  & IN-21K & 30.2 & 31.7 & 14.9 & - & - & - & - & - & - \\
         DCR~\cite{dcr} & & IN-1K & 31.0 & 31.1 & 14.6 & - & - & - & - & - & - \\
         TeSTra~\cite{testra} &  & IN-1K & 26.8 & 36.2 & 17.0 & 27.1 & 30.1 & \textbf{13.3} & 19.3 & 28.6 & 13.7 \\
         MAT (Ours) &  & IN-1K & \textbf{32.7} & \textbf{39.7} & \textbf{18.8} & \textbf{31.7} & \textbf{32.1} & 12.7 & \textbf{25.7} & \textbf{32.4} & \textbf{16.0} \\
         \hline
         RULSTM~\cite{rulstm} & \multirow{4}{*}{{RGB + OF + OBJ}} & IN-1K & 27.8 & 30.8 & 14.0 & 28.8 & 27.2 & 14.2 & 19.8 & 22.0 & 11.1 \\
         TempAgg~\cite{tempagg} &  & IN-1K & 23.2 & 31.4 & 14.7 & 28.0 & 26.2 & \textbf{14.5} & 14.5 & 22.5 & 14.8 \\
         AVT+~\cite{avt} &  & IN-1K & 25.5 & 31.8 & 14.8 & 25.5 & 23.6 & 11.5 & 18.5 & 25.8 & 12.6 \\
         AVT+~\cite{avt} &  & IN-21K & 28.2 & 32.0 & 15.9 & 29.5 & 26.0 & 12.8 & 23.2 & 29.2 & 14.1 \\
        \hline
         TeSTra~\cite{testra} & \multirow{2}{*}{RGB + OF} & IN-1K & 30.8 & 35.8 & 17.6 & 29.6 & 26.0 & 12.8 & 23.2 & 29.2 & 14.2 \\
        MAT (Ours) & & IN-1K & \textbf{35.0} & \textbf{38.8} & \textbf{19.5} & \textbf{32.5} & \textbf{30.3} & 13.8 & \textbf{28.7} & \textbf{33.1} & \textbf{16.9}  \\
    \arrayrulecolor{black}
    \end{tabular}
    }
    \caption{\textbf{Comparison to prior work on EPIC-Kitchens-100 Action Anticipation~\cite{ek100}.} 
Accuracy measured by class-mean recall@5(\%) following the standard protocol.}
    \vspace{-3mm}
    \label{tab:ek100_anticipation}
\end{table*}

\begin{table}[t]
    \centering
    \small
    \setlength{\tabcolsep}{1.2mm}{
    \begin{tabular}{lcccccc}
         Modality & Encoder & Init & FT & Overall & Unseen & Tail \\
         \midrule
    RGB & TSN & IN-1K & \checkmark & 18.8 & 12.7 & 16.0 \\
    RGB + OF & TSN & IN-1K & \checkmark & 19.5 & 13.8 & 16.9 \\
    RGB & ViT-L & K400 & & 19.1 & 16.2 & 16.1 \\
    RGB (V) & ViT-L & Ego4D & & 21.9 & 19.4 & 18.5 \\
    RGB (N) & ViT-L & Ego4D & & 24.2 & 21.4 & 20.9 \\
    RGB (V + N) & ViT-L & Ego4D & & 24.6 & 21.4 & 21.2 \\
    \arrayrulecolor{black}
    \end{tabular}
    }
    \caption{\textbf{Exploring different visual encoders} for EK100 action anticipation. V or N denotes the encoder is pretrained on verb or noun subset~\cite{internvideo}. ``FT'' is that the encoder is fine-tuned on the classification task of EK100.}
    \label{tab:pretrain}
    \vspace{-5mm}
\end{table}

\textbf{Future Supervision.}
In addition, we carry out diverse experiments to determine the optimal choice of future supervision, as illustrated in Table~\ref{tab:future supervision}. We investigate two methods, namely future feature supervision and future label supervision. The results suggest that, compared to feature supervision, label supervision yields superior accuracy by incorporating stronger semantics derived from manual annotation.
Furthermore, we explore the significance of deep supervision, \ie, multi-stage supervision. We observe that introducing multiple supervisions of real action labels improves performance.

\textbf{Shared Classifier.}
In Table~\ref{tab:classifier}, we exploit the classifier design in deep supervision for short-term memory and anticipation. 
We observe that sharing classifier parameters separately for them yields better performance. When we use a full-shared classifier for memory and anticipation, sharing all feature samples between them leads to the best performance. These results indicate that all yielded features in the interaction processes enrich the feature samples to be classified, reaching a function similar to the data augmentation.

\begin{table}
    \centering
    \small
    \setlength{\tabcolsep}{2.5mm}{
    \begin{tabular}{lcccc}
         \multirow{2}{*}{Method}   & \multicolumn{2}{c}{THUMOS'14} & \multicolumn{2}{c}{TVSeries} \\
        \cmidrule(lr){2-3} \cmidrule(lr){4-5}
        & Kinetics & ANet & Kinetics & ANet  \\
        \midrule
        RED~\cite{red} & - & 37.5 & 75.1 & - \\
        TRN~\cite{trn} & - & 38.9 & 75.7 & - \\
        OadTR~\cite{oadtr} & 53.5 & 45.9 & 77.8 & 79.1 \\
        LSTR~\cite{lstr} & 52.6 & 50.1 & 80.8 & - \\
        GateHUB~\cite{gatehub} & - & 54.2 & 82.0 & - \\
        TesTra~\cite{testra} & 56.8 & 55.3 & - & - \\
        \rowcolor{light}
        MAT (ours) & \textbf{58.2} & \textbf{57.3} & \textbf{82.6} & \textbf{81.5} \\
    \end{tabular}}
    \caption{\textbf{Action anticipation result} on THUMOS'14 and TVSeries, mAP is reported for THUMOS'14 and mcAP for TVSeries.}
    \vspace{-3mm}
    \label{tab:anticipation}
\end{table}

\textbf{Conditional Circular Interaction.}
Table~\ref{tab: interaction} compares conditional circular interaction incorporating different interaction information streams, methods, and times. From (a) to (e), our experimental results demonstrate that the model's classification performance improves with the increasing richness of interaction information.
Moreover, we observe that latent information in the future is equally or even more critical to classify the current action than information from the distant past, as evidenced by the performance comparison between (c) and (e).
We also investigate the effect of different interaction mechanisms, comparing (f) and (g), and the results indicate that the cross-attention (CA) mechanism outperforms average pooling and concatenation similar to~\cite{oadtr}, achieving a 2.1\% higher accuracy.
Lastly, we evaluate the impact of different interaction times $N_t$, as illustrated by (g), (h), and (i). The experimental results reveal that the model's performance improves gradually with the increase of $N_t$. Notably, the model attains its highest performance when interacting with all information $N_t=2$ times.

\begin{table*}[t]
    \centering
    \small
    \setlength{\tabcolsep}{3mm}{
    \begin{tabular}{lccccccccccccc}
        \multirow{2}{*}{Method} & \multicolumn{2}{c}{Augmentation} & \multicolumn{3}{c}{Overall} & \multicolumn{3}{c}{Unseen} & \multicolumn{3}{c}{Tail}  \\
        \cmidrule(lr){2-3} \cmidrule(lr){4-6} \cmidrule(lr){7-9} \cmidrule{10-12}
        & Long & Short & Verb & Noun & Action & Verb & Noun & Action & Verb & Noun & Action \\
        \hline
        LSTR$^\ddagger$~\cite{lstr} & - & - & 39.6 & 44.1 & 22.6 & 34.3 & 35.3 & 18.7 & 39.0 & 41.6 & 20.7  \\
        TesTra$^\ddagger$~\cite{testra} & - & - & 40.0 & 44.8 & 23.2 & 34.6 & 36.0 & 19.0 & 39.4 & 42.1 & 20.9 \\
        \rowcolor{light}
        MAT~(ours) & - & - & \textbf{41.8} & \textbf{46.1} & \textbf{24.9} & \textbf{36.5} & \textbf{37.9} & \textbf{20.1} & \textbf{40.8} & \textbf{43.2} & \textbf{22.8} \\
        \hline
        TesTra$^\ddagger$~\cite{testra} & MixClip & - & 39.7 & 45.6 & 25.1 & 36.3 & 37.2 & 19.2 & 39.0 & 42.4 & 22.2 \\
        MAT~(ours) & MixClip & - & 42.6 & 47.3 & 25.9 & 38.3 & 37.6 & 19.7 & 41.8 & 44.0 & 23.1 \\
        \rowcolor{light}
        MAT~(ours) & MixClip & MixClip+ & \textbf{44.5} & \textbf{48.3} & \textbf{26.3} & \textbf{39.9} & \textbf{38.1} & \textbf{20.3} & \textbf{43.4} & \textbf{46.6} & \textbf{23.7} \\
    \end{tabular}}
    \caption{\textbf{Online action detection result} on EPIC-Kitchens-100. Accuracy is measured by class-mean recall@5(\%). $^\ddagger$ was reproduced by us because LSTR~\cite{lstr} and TesTra~\cite{testra} did not report the result.}
    \vspace{-3mm}
    \label{tab:oad_ek100}
\end{table*}

\begin{table}[t]
    \centering
    \small
    \setlength{\tabcolsep}{1mm}{
    \begin{tabular}{lcccccc}
        \multirow{2}{*}{Method}  & \multirow{2}{*}{Arch} & \multicolumn{2}{c}{THUMOS'14} & \multicolumn{2}{c}{TVSeries} & HDD \\
        \cmidrule(lr){3-4} \cmidrule(lr){5-6} \cmidrule(lr){7-7}
        & & Kinetics & ANet & Kinetics & ANet & Sensor \\
        \midrule
        CDC~\cite{cdc}        & CNN & - & 44.4 & - & - & - \\
        RED~\cite{red}        & RNN & - & 45.3 & - & 79.2 & 27.4 \\
        TRN~\cite{trn}        & RNN & 62.1 & 47.2 & 86.2 & 83.7 & 29.2 \\
        IDN~\cite{idn}       & RNN & 60.3 & 50.0 & 86.1 & 84.7 & - \\
        LAP~\cite{LAP-Net}    & RNN & - & 53.3 & - & 85.3 & - \\
        PKD~\cite{pkd}        & CNN & 64.5 & - & 86.4 & - & - \\
        WOAD~\cite{woad}     & RNN & 67.1 & - & - & - & - \\
        OadTR~\cite{oadtr}    & Trans & 65.2 & 58.3 & 87.2 & 85.4 & 29.8 \\
        Colar~\cite{colar}    & Trans & 66.9 & 59.4 & 88.1 & 86.0 & 30.6 \\
        LSTR~\cite{lstr}      & Trans & 69.5 & 65.3 & 89.1 & 88.1 & - \\
        GateHUB~\cite{gatehub}  & Trans & 70.7 & 69.1 & 89.6 & 88.4 & 32.1 \\
        TeSTra~\cite{testra}  & Trans & 71.2 & 68.2 & - & - & - \\
        \rowcolor{light}
        MAT (Ours)  & Trans & \textbf{71.6} & \textbf{70.4} & \textbf{89.7} & \textbf{88.6} & \textbf{32.7} \\
    \end{tabular}
    }
    \caption{\textbf{Online action detection performances} on THUMOS'14~\cite{THUMOS14}, TVSeries~\cite{oad2016}, and HDD~\cite{hdd}. The mAP performance is reported for THUMOS'14 and HDD, while the mcAP is reported for TVSeries.}
    \vspace{-3mm}
    \label{tab:oad result}
\end{table}

\textbf{MixClip+ for Anticipation.}
Table~\ref{tab:mc+} presents the augmentation efficacy of MixClip+ or MixClip on long- and short-term memory. Notably, the baseline experiences a significant drop in action recall to 18.1\% without data augmentation. However, the result is still significantly better than Testra~\cite{testra} with MixClip (18.1\% \emph{vs} 17.6\%), as shown in Table~\ref{tab:ek100_anticipation}.
Furthermore, the results in row 3 reveal that MixClip is not suited for short-term memory, as it cannot improve performance and may even degrade the results regarding the verb (32.1\% \emph{vs} 31.0\%). However, the model achieves the best performance (19.5\%) when performing MixClip and MixClip+ augmentation separately on long- and short-term memory. Moreover, we validate the augmentation of double-MixClip+ (in row 5) on long- and short-term memory and observe that the overall gain of the model is small, similar to double-MixClip (in row 3).

\subsection{Comparison with State-of-the-Art Methods}

\textbf{Action Anticipation.}
We compare MAT with prior methods on EPIC-Kitchens-100 for action anticipation, as shown in Table~\ref{tab:ek100_anticipation}. We divide the experimental results into two parts, one part of the input modality is RGB, and the other is multi-modal input, such as optical flow and object features.
Using the ImageNet-1K pretrained features, MAT significantly outperforms RULSTM~\cite{rulstm}, AVT~\cite{avt}, and TeSTra~\cite{testra} on terms of verb, noun, and action (at least 2.5\%,  3.0\%, and 1.8\%, respectively). MAT with RGB and optical flow features achieves 1.8\% higher action recall than TeSTra with the same input, which can better reflect the effectiveness of the modules we designed.

In Table~\ref{tab:pretrain}, we conduct an extensive analysis of the impact of various visual encoders on the performance of action anticipation. In particular, we leverage the ViT architecture~\cite{vit}, which is pre-trained on the largest egocentric dataset Ego4D \cite{ego4d}, made available by~\cite{internvideo}, owing to its superior performance in Ego4D Challenges. Additionally, to compare the impact of first-person and third-person pretraining, we utilize the weights pretrained on the K400~\cite{kinetics}, provided by VideoMAE \cite{videomae}. 
The results suggest that encoders pre-trained on first-person datasets perform better than on widely-used third-person datasets. Moreover, leveraging the noun subset brings a huge performance boost, comparing pretraining on the verb subset of the Ego4D dataset. We suppose the noun subset with more object categories (the verb and noun subset include 118 and 582 categories) encourages the encoder to search more complex visual semantics.
It highlights the critical role of egocentric perspective upstream pretraining method~\cite{egovlp, lavila} and diverse semantic annotations~\cite{ego4d, ek100, egotaskqa}, which could have far-reaching implications for egocentric research.

Following previous works~\cite{lstr, gatehub, testra}, we further evaluate MAT on action anticipation tasks for THUMOS'14 and TVSeries. Unlike the previous work that required additional learnable tokens, we take the corresponding tokens from $ \mathbf{F}_A $ for forecasting. Table~\ref{tab:anticipation} shows that using the ActivityNet pretrained features, MAT significantly outperforms the existing methods by 2.0\% and 0.6\% on THUMOS'14 and TVSeries, respectively.  Moreover, the performance of MAT can also be further improved when pretraining on Kinetics.

\textbf{Online Action Detection.}
We also compare MAT with other state-of-the-art online action detection methods~\cite{trn, lstr, oadtr, colar, gatehub} on THUMOS'14, TVSeries, and HDD. As illustrated in Table \ref{tab:oad result}, our MAT achieves state-of-the-art performance and improves mAP by 2.2\% on the THUMOS'14 dataset under the TSN-Anet feature input, and 0.4\% under the TSN-Kinetics feature. For TVSeries, MAT is slightly better than GateHUB by 0.1\% and 0.2\% on features from Kinetics and ActivityNet, respectively. Compared to other dataset, we believe that the metrics on TVSeries have been saturated.
Furthermore, MAT achieves 0.6\% better than the state-of-the-art GateHUB~(32.1\%) method on HDD. 

In Table~\ref{tab:oad_ek100}, we offer the results for online action detection on the EK100 dataset. We divide the experimental results into
two parts: one does not use augmentation, and the other uses MixClip or MixClip+. The results show that MAT surpasses the previous method by a large margin whether or not to use data augmentation. It is worth noting that when performing online action detection, we only need to take out the corresponding token in the short-term memory $\mathbf{\widehat{M}}_S $ for prediction. The superior Performance of the model further highlights that MAT can simultaneously handle online action detection and anticipation tasks, thus integrating the two problems into a unified framework and performing excellently.

\begin{table}[t]
    \centering
    \small
    \setlength{\tabcolsep}{0.9mm}{
        \begin{tabular}{lccccccc}
        \multirow{2}{*}{Method} &  \multirow{2}{*}{\#Param}  & \multicolumn{5}{c}{FPS} & \multirow{2}{*}{mAP} \\
        \cmidrule(lr){ 3 - 7 }& &  \begin{tabular}[c]{@{}c@{}}
        Optical \\ Flow
        \end{tabular} & \begin{tabular}[c]{@{}c@{}}
        RGB \\ Feat
        \end{tabular}  & \begin{tabular}[c]{@{}c@{}}
        Flow \\ Feat
        \end{tabular}   & Model & Total \\
        \hline TRN  & 402.9 M & 8.1 & 70.5 & 14.6 & \textbf{123.3} & 8.1 & 47.2 \\
        OadTR & 75.8 M & 8.1 & 70.5 & 14.6 & 110.0 & 8.1 & 58.3 \\
        LSTR  & 58.0 M & 8.1 & 70.5 & 14.6 & 91.6 & 8.1 & 65.3 \\
        GateHUB & 45.2 M & 8.1 & 70.5 & 14.6 & 71.2 & 8.1 & {69.1} \\
        \hline 
        MAT (Ours) & 94.6 M & 8.1 & 70.5 & 14.6 & 72.6 & 8.1 & \textbf{70.4} \\
        \end{tabular}
        }
    \caption{\textbf{Efficiency comparison} between MAT and the previous work on parameter (M) and inference speed (FPS)}
    \label{tab:speed}
\end{table}

\begin{figure}[t]
    \centering
    \includegraphics[width=0.48\textwidth]{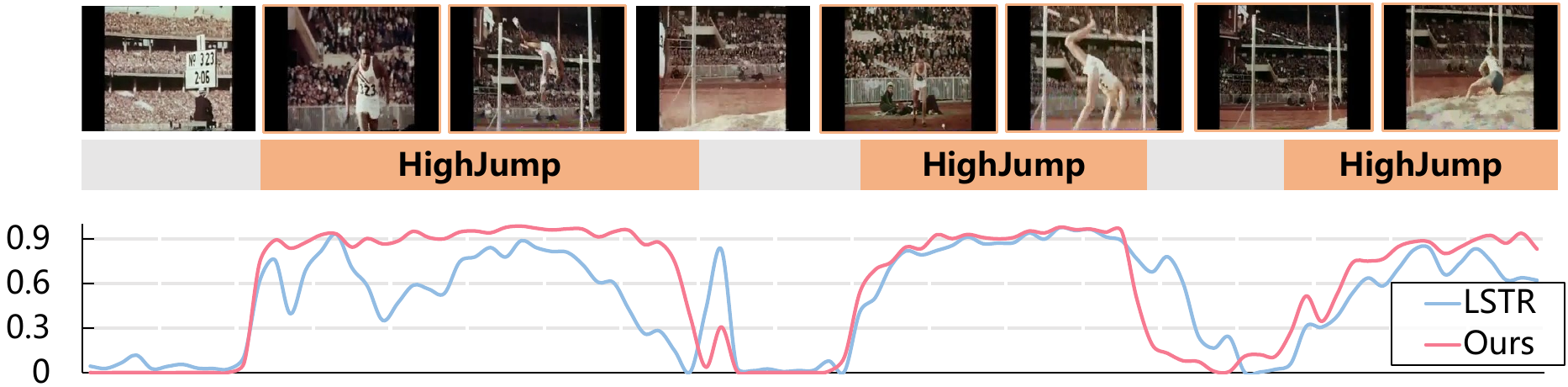}
    \caption{\textbf{Visualization of MAT's online prediction.} The curves indicate the predicted confidence of the ground-truth class (\emph{HighJump}) with LSTR and  our method.}
    \vspace{-3mm}
    \label{fig:vis}
\end{figure}

\subsection{Efficiency Analysis}
Table \ref{tab:speed} compares MAT with other methods in terms of model parameters and running time, conducted on a single Tesla V100. It is worth noting that under the same features, the efficiency bottleneck of the system remains in optical flow calculation and feature extraction. Our approach achieves a stronger performance while effectively balancing the trade-off between parameters and computational.

\subsection{Qualitative Analysis.}
In Fig \ref{fig:vis}, we qualitatively analyzes the proposed MAT model. The confidence in the range $[0, 1]$ on the y-axis denotes the probability of predicting the current action,\eg, \emph{HighJump}. The figure highlights our MAT's successful suppression of background and high-confidence feedback of action segments compared with LSTR. The circular interaction of memory and anticipation provides reliable semantic information for the online prediction of the model, making the model offers a more sensitive and accurate torsion signal on the transition between background and foreground.

\begin{figure}
    \centering
    \includegraphics[width=0.48\textwidth]{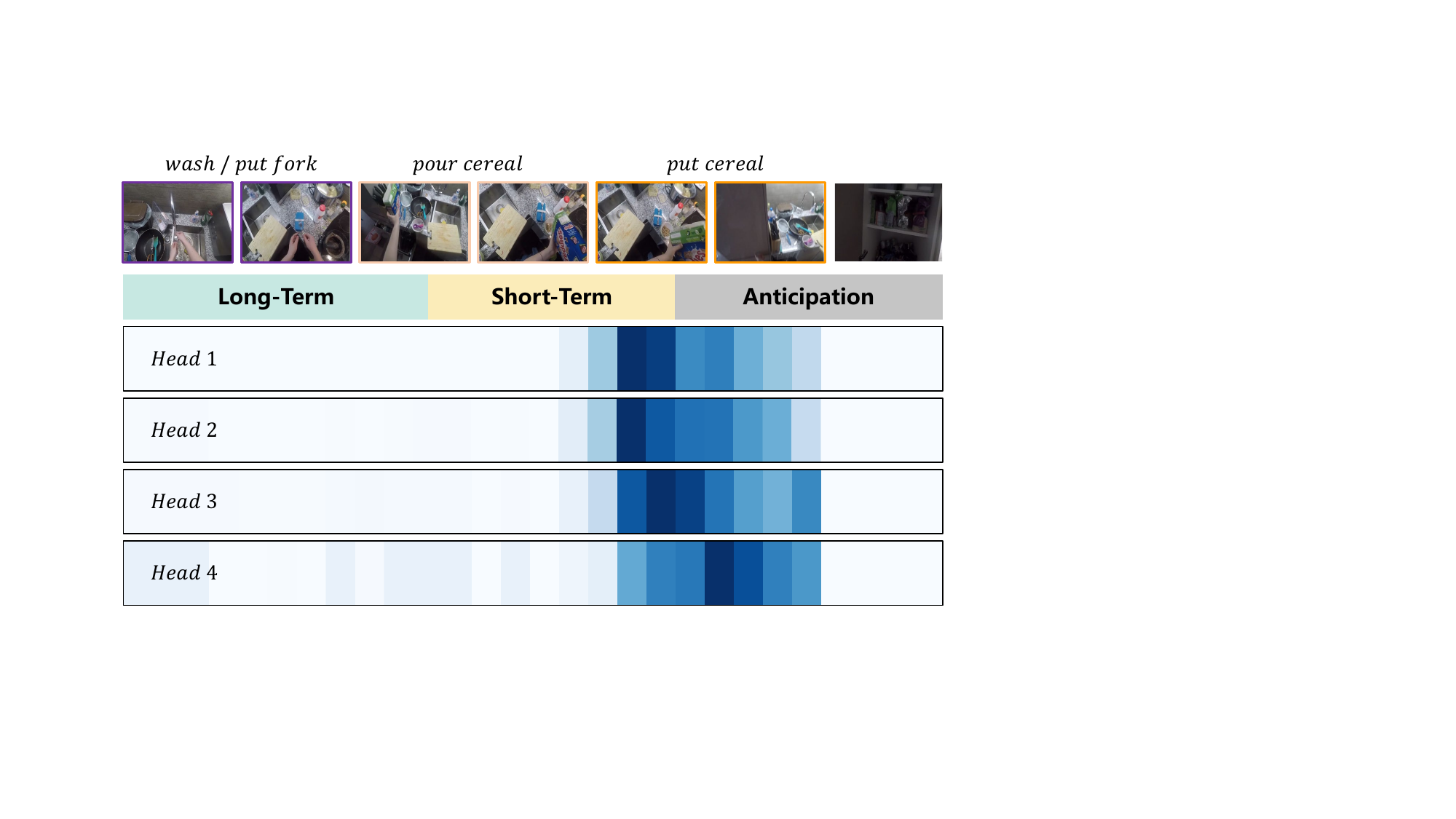}
    \caption{\textbf{Attention weight visualization of different attention heads.}}
    \vspace{-3mm}
    \label{attn_vis}
\end{figure}

\subsection{Attention Visualization}
Fig~\ref{attn_vis} depicts the attention weights of the last token of the short-term memory. It can be seen that under the blessing of future perception, MAT's multi-head attention mechanism pays attention to memory and anticipation simultaneously, leading to the model not limiting in history and attaining better performance.

\vspace{-0.5em}
\section{Conclusion and Future Work}
\label{sec:conclusion}
We present Memory-and-Anticipation Transformer (MAT), a novel memory-anticipation-based paradigm for online action detection and anticipation, to overcome the weakness of most existing methods that can only complete modeling temporal dependency within a limited historical context. 
Through extensive experiments on four challenging benchmarks across two tasks, we show its applicability in predicting present or future actions, obtaining state-of-the-art results, and demonstrating the importance of circular interaction between memory and anticipation in the entire temporal structure. Although the temporal dimension is only considered in this work, 
we believe MAT would be a general paradigm for any AI system used to analyze video online and predict future events.
In the near future, we will validate our model on more benchmarks and extend it to the long-term anticipation task. In the longer term, we will continue to unearth the intrinsic association of memory with anticipation and develop more effective forecasting models.

{\small
\bibliographystyle{ieee_fullname}
\bibliography{egbib}
}

\end{document}